\documentclass{article}

\usepackage{neurips_wrl2023}

\usepackage[utf8]{inputenc} 
\usepackage[T1]{fontenc}    
\usepackage{hyperref}       
\usepackage{url}            
\usepackage{booktabs}       
\usepackage{amsfonts}       
\usepackage{nicefrac}       
\usepackage{microtype}      
\usepackage{fancyvrb} 
\usepackage{multirow}
\usepackage{makecell}
\usepackage{graphicx} 
\usepackage{natbib}
\setcitestyle{authoryear,open={((},close={))}} 

\title{Exploring and Improving the Spatial Reasoning Abilities of Large Language Models }

\author{%
  David S.~Hippocampus\thanks{Use footnote for providing further information
    about author (webpage, alternative address)---\emph{not} for acknowledging
    funding agencies.} \\
  Department of Computer Science\\
  Cranberry-Lemon University\\
  Pittsburgh, PA 15213 \\
  \texttt{hippo@cs.cranberry-lemon.edu} \\
}

\begin{document}

\maketitle

\begin{abstract}
Large Language Models (LLMs) represent formidable tools for sequence modeling, boasting an innate capacity for general pattern recognition. Nevertheless, their broader spatial reasoning capabilities, especially applied to numerical trajectory data, remain insufficiently explored. In this paper, we investigate the out-of-the-box performance of ChatGPT-3.5, ChatGPT-4 and Llama 2 7B models when confronted with 3D robotic trajectory data from the CALVIN baseline and associated tasks, including 2D directional and shape labeling. Additionally, we introduce a novel prefix-based prompting mechanism, which yields a 33\% improvement on the 3D trajectory data and an increase of up to 10\% on SpartQA tasks over zero-shot prompting (with gains for other prompting types as well). The experimentation with 3D trajectory data offers an intriguing glimpse into the manner in which LLMs engage with numerical and spatial information, thus laying a solid foundation for the identification of target areas for future enhancements.

\end{abstract}

\section{Introduction}
Large Language Models (LLMs), e.g. GPT-4 [\citenum{openai2023gpt4}] \& PaLM [\citenum{chowdhery2022palm}], are massive models trained on diverse corpora with billions of tokens of text. Recent works have established the competence of LLMs in extrapolating more abstract, non-linguistic patterns, thus allowing them to serve as "general pattern machines [\citenum{mirchandani2023large}]. As such, in addition to the text-based tasks for which they were trained, LLMs successfully demonstrate cross-disciplinary capabilities, such as high-level planning for robotic policies [\citenum{huang2022language}, \citenum{xie2023translating}, \citenum{ahn2022i}], reward function design [\citenum{kwon2023reward}, \citenum{hu2023language}], and math \& logic puzzles.

Labeling of various kinds of data [\citenum{he2023annollm}] falls under the paradigm of general pattern matching, and is of utmost practical use in the prospective organization of unlabeled raw datasets. One such application is in the space of language \& robotics; there are a limited number of datasets that supply language annotations for each demonstration (e.g. the trajectory is described by the instruction "pick up the blue cup"), as human labeling is costly. Intrigued by the potential for LLMs in annotating large-scale robotic datasets, we investigate LLM labeling as applied to 2D and 3D robotic trajectory data, i.e. the ability to describe a sequence of n-dimensional points with the type of motion it embodies, such as "lifting". While it is possible to explicitly train models for these capabilities, this work instead focuses on the inherent abilities of LLMs out-of-the-box, which may have downstream implications for broader numerical and spatial queries, e.g. trend analysis or time-series data.

Additionally, considering that LLMs appear to struggle with spatial reasoning abilities [\citenum{bang2023multitask}, \citenum{xie2023translating}, \citenum{cohn2023dialectical}] (as defined by an understanding of shapes and relationships between different objects and spaces), we also gauge whether there are unique factors about this kind of numerical \& spatial data that impact the improvements furnished by prompting techniques like In-context Learning (ICL) and Chain-of-Thought (CoT) prompting. Subsequently, we propose a strategy to pre-fix a prompt with a more general example to achieve greater performance gains for this type of application.

To summarize, we are overall interested in empirically answering the following research questions:
\textbf{RQ1} Can LLMs be used to identify simple spatial patterns (e.g. circles or straightforward directions)?\\
\textbf{RQ2} Can LLMs be used to identify and label more complex spatial patterns (such as more irregular 3D trajectories)? As such, does the irregularity of the pattern make it difficult to CoT-type reasoning-based prompting?\\
\textbf{RQ3} Is there any knowledge transfer that can happen from simpler spatial tasks to more complex ones that can impact overall performance? 

\begin{figure}
  \centering
  \includegraphics[width=14cm]{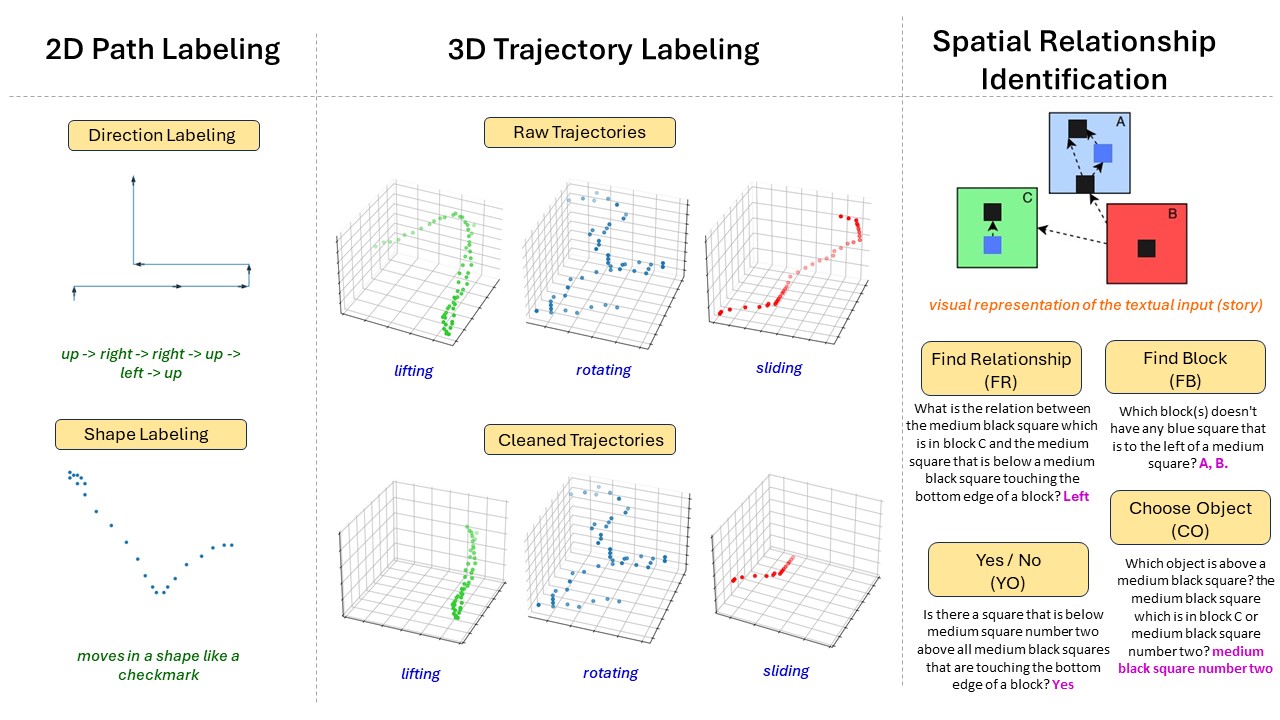}
  \caption{Illustrations of the Various Spatial Tasks: 2D Path Labeling of both directions (e.g. "up", "left", etc.) and shapes, 3D Trajectory Labeling of "lifting", "rotating" and "sliding" motions (as well as the cleaned versions), and relationship identification between blocks in an imagined setup.}
  \label{fig:tasks}
\end{figure}

\section{Related works}
\textbf{LLMs and Spatial Pattern Matching.} Previous work has shown that LLMs can improve and complete low-level robotic action sequences or repetitive progressions like a sinusoid [\citenum{mirchandani2023large}]. However, labeling of a sequence in its entirety has not been addressed, which requires long-form context retention, semantic comprehension to link a sequence to its textual annotation, and generalization to non-repetitive, complex patterns. Some works show that LLMs acceptably label one-dimensional time-series data [\citenum{xue2023promptcast}, \citenum{liu2023large}], but assess much shorter sequences, exclude higher dimensional analysis, or use additional token embedding models. Furthermore, on the whole, prevailing examinations of LLMs' spatial reasoning abilities [\citenum{bang2023multitask}, \citenum{xie2023translating}, \citenum{cohn2023dialectical}] have revealed a considerably poor performance. We extend these works by tackling the inherent performance of LLMs on the underexplored subproblem of higher-dimensional trajectory identification.

\textbf{LLMs \& 3D Robotics. } LLMs have been applied across a number of areas in robotics, most recently in originating high level step-by-step plans from task descriptions [\citenum{ahn2022i}, \citenum{huang2022language}] and robot policy code publication [10, 11]. However, our work falls into the bucket of whether LLMs can directly understand control (e.g., at the level of trajectories) in a zero-shot manner, which remains an open problem.
There are also works in the embodied robotics domain where LLMs are used to reason about a 3D point-cloud scene [\citenum{hong20233dllm}, \citenum{takmaz2023openmask3d}, \citenum{runsen2023pointllm}], but these papers either use a vision model (or joint vision-language model) to integrate visual embeddings or append a finite number of 3D object positions acquired using a detection model to the prompt. Our approach is distinguished by focusing on understanding continuous sequences of 3D points on the prompt-side.

\textbf{LLMs \& Prompting}
Several prompting approaches have been shown to improve results, such as In-context Learning [\citenum{brown2020language}], which supplies few-shot examples that guide the model, and Chain-of-Thought [\citenum{wei2023chainofthought}], which harnesses the ability of LLMs to adhere to a guided thought process for problem solving. The presence of symbols, patterns and texts are crucial to the effectiveness of CoT and ICL [\citenum{madaan2022text}], but whether spatial trends follow such an archetype has yet to be explicitly examined. 

\section{Language Models as Trajectory Labelers}

\subsection{2D Path Labeling}
\subsubsection{Direction Labeling}
As discussed in [\citenum{mirchandani2023large}] the ability of an LLM to pattern match is driven by in-context learning on the provided numerical tokens, which can be formulated as the problem of using the context $s_{1:k} =(s_1, ..., s_k)$, where each $s_i$ is a symbol and using it to autoregressively predict $s_{k+1}$ by using the factorized conditional probability $p(s_{1:k}) = \Pi_{i=1}^n p(s_i | s_1, ..., s_{i-1})$. Usual in-context learning examples segment the prompt into continuations of multiple examples, each a variable length sequence: $x_{1:k} =(x_1, ..., x_k)$ where each $x_i = (s_1^i, s_2^i, ..., s_{m^i}^i)$.

We adapt this paradigm for directional labeling by having $x_1$ state the model's expertise in spatial analysis and prompt it to generate direction labels for a newly provided sequence given the examples. Then each $x_i$ from $i=2$ onwards is an example, an input-output pair $D_i; L_i$ where $D_i$ is the sequence of symbol aggregates $(d_1^i, d_2^i, ... d_{j}^i)$ of length $j$, with each $d_k^i$ further being separated out into the symbols $(d_{k_x}^i, d_{k_y}^i)$, representing a coordinate in 2D Cartesian space. $L_i$ is similarly a sequence of words $(l_1^i, l_2^i, ... l_{j-1}^i)$ of length $j-1$, where each $l_k^i$ is a word representing the direction that describes the movement from $d_k^i$ to $d_{k+1}^i$ and is one of 
\Verb"[left, right, up, down]". There are thus $j$ points and $j-1$ segment labels (see Fig. ~\ref{fig:tasks}).

\subsubsection{Shape Identification}
Using a similar prompt framework to the one described above, $x_1$ states the model's expertise in spatial analysis and prompts it to identify the overall shape of the movement represented by the list of 2D coordinates; for example, "moves along a path that mirrors the pattern of a checkmark" or "moves in a circular path". Since the dataset size is fairly limited, no in-context learning is applied and the model is evaluated zero-shot by specifying $x_2$ as the input sequence of $(x, y)$ coordinates $(d_1^i, d_2^i, ... d_{j}^i)$ (see Fig. ~\ref{fig:tasks}).

\subsection{3D Trajectory Labeling}

\subsubsection{Zero-shot Prompting}
To provide a baseline for the various prompting mechanisms probed, we initially implement zero-shot prompting, in which no exemplars are imparted to the model. Therefore, analogous to 2D experiments, $x_1$ states the model's expertise in spatial analysis and prompts it to classify the type of motion exemplified by the sequence into one of $N$ categories, but there are no further sequences (see Fig. ~\ref{fig:prompts}). For our set of experiments, we designate three classes that correspond to meaningfully and spatially disparate motions - lifting, rotating and sliding.

\subsubsection{In-context Learning}
\begin{figure}
  \centering
  \includegraphics[width=14cm]{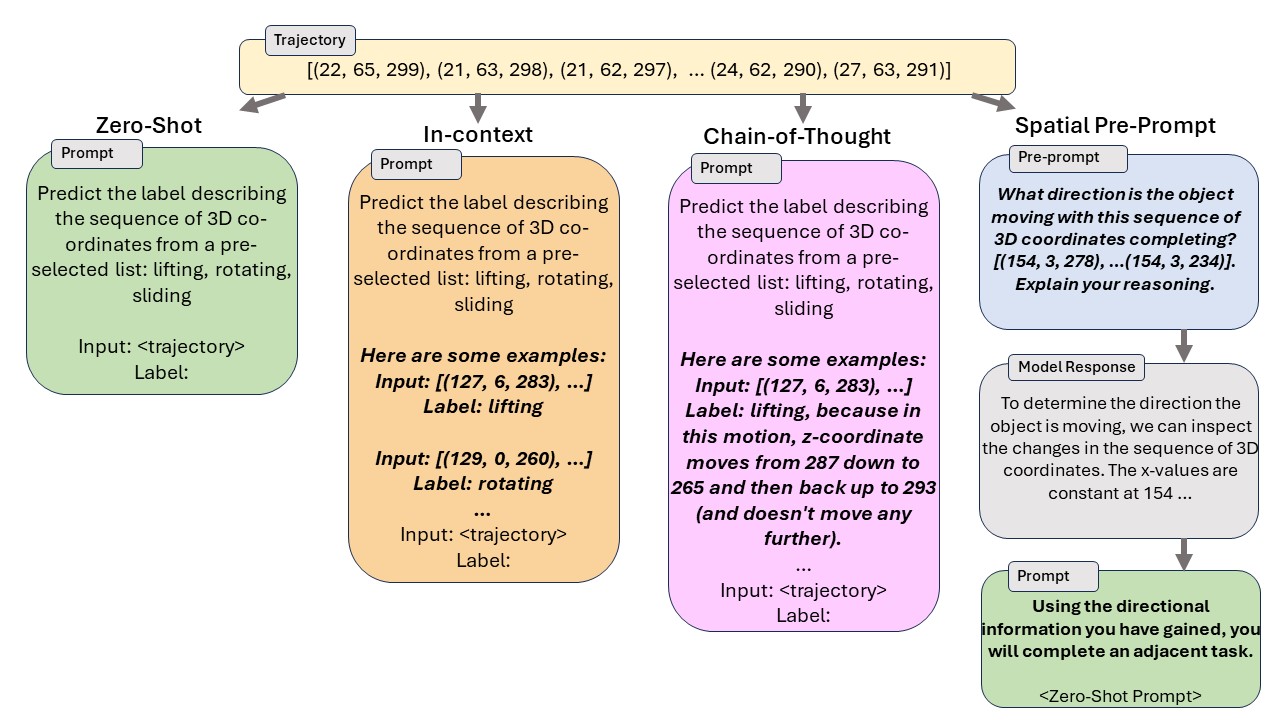}
  \caption{Different Types of Prompting Mechanisms - Zero-shot, In-context Learning, Chain-of-Thought and Spatial Prefix-Prompting. In Spatial Pre-Prompt, a tangential question is first asked, to which the model provides a response, following which the primary query is inquired.}
  \label{fig:prompts}  
\end{figure}

In-context Learning (ICL) [1] supplies a few samples that the model can generalize from. $x_1$ declares that the model that it is an expert in 3D trajectory labeling and prompts the model to classify the type of motion exemplified by the sequence into one of $N$ categories. Then each $x_i$ from $i=2$ onwards is an input-output pair $D_i; l_i$ where $D_i$ is the sequence of symbol aggregates $(d_1^i, d_2^i, ... d_{j}^i)$, with each one representing a coordinate in 3D Cartesian space $(d_{k_x}^i, d_{k_y}^i), d_{k_z}^i)$. $l_i$ is a single word describing the motion and belongs to one of the $N$ classes (see Fig. ~\ref{fig:prompts}).

\subsubsection{Chain-of-Thought Prompting}
Chain-of-Thought Prompting (CoT) [\citenum{wei2023chainofthought}], in which the few-shot examples are augmented by a step-by-step reasoning, has superseded In-context Learning on an array of textual and mathematical reasoning tasks. We extend it to our task as follows. 
The starting guideline for $x_1$ is the same as in-context learning, with the declaration that the model is an expert in 3D trajectory labeling and a prompt for the model to classify the type of motion exemplified by the sequence into one of $N$ categories. However in CoT, successive pairs $x_2, x_3, ..., x_{2t}, x_{2t+1}$ for some $t$, represent the few-shot examples. The even $x_{2t}$ is the representative sequence of 3D coordinates and the odd $x_{2t+1}$ is its associated answer, which is the motion-type label and accompanying reasoning steps (such as the fact that back-and-forth changes in the $x$ and $y$ coordinates can hint at a rotating motion, see Fig. ~\ref{fig:prompts}).

\subsubsection{Spatial Prefix-Prompting}
Anticipating the challenge with generalizing from irregular examples and bolstered by the model's performance on simpler 2D data, we also propose a new method of prompting called "Spatial Prefix-Prompting" (SPP). The method draws from a prior selection of fixed questions that instigate the model to first ponder a tangentially related spatial problem (e.g. identifying the single direction an object is moving in or checking whether a point is in the center of a circle), and then use the "knowledge gained" to answer an adjacent question, such as labeling a new, more complex 3D trajectory (see Fig. ~\ref{fig:prompts}). This technique does not necessitate the more intensive CoT-style curation of step-by-step examples, and we hypothesize that it may build upon the more fundamental spatial concepts a model is trained on to generalize better than few-shot learning (wherein the selected examples may not be representative of all the trajectories).


\section{Experiments}

\subsection{Implementation Details}
\subsubsection{2D Labeling and Description}
There aren't many datasets for elementary 2D shapes and directions, and given the ease of generating such data, we decided to autogenerate the datasets. 
For the direction labeling task, a dataset of size 30 is generated with 10 short-horizon sequences of 2D coordinates (of length 6-8 segments), 10 long-horizon long sequences (length 35 - 40), and 10 short floating-point sequences using Python's NumPy package, with each segment's size and the directions randomly chosen based on a fixed seed. 
We experiment with both 1) scaling the values to integers between [0, 100] (to use fewer tokens to represent a single number) 2) scaling the values to double-digit fractions between [0, 100] (to test whether the higher token representation translates to higher precision). Only Zero-shot and In-context Learning are applied for Shape and Direction Labeling respectively. For the shape identification, we use some previously collected hand-gesture data in which human demonstrators sign a variety of shapes, including circles and check marks, and 2D positions of the finger are recorded; the dataset tests the model on the inherent noise from human demonstrations. A subset of the dataset is "cleaned" for use by having an expert to remove extraneous points that don't belong to the shape, and it is normalized to the range [0, 100] (both integer and floating point). The size of the dataset is 13.

\subsubsection{3D Trajectory Labeling}
We use the CALVIN benchmark [\citenum{mees2022calvin}], a dataset for learning long-horizon language-conditioned tasks for robotics. It includes 3D end-effector positions (can be extracted from the low-dimensional state vector) and associated language descriptions of the action the robot is attempting to complete. Due to time and resource constraints from the human evaluations, we select only a small subset of the CALVIN dataset (30 samples) and three disparate subtasks ("rotate", "lift", "slide"). The trajectories are often complex and may not intuitively always resemble the action being completed, with many extraneous movements (see Fig. ~\ref{fig:tasks}).  Therefore, we also create a version of this dataset called "CALVIN-Cleaned" in which a human annotator extracts the parts of the trajectory that match with the specific action (e.g. only the lifting portion, see Fig. ~\ref{fig:tasks}). Note, the CALVIN-Cleaned dataset retains the original "rotate" trajectories, as the task description is linked to the back-and-forth changes in the entirety of the motion, not any particular subsection. Finally, the dataset is normalized to the range $[0, 300] \in \mathbb{Z}$ to increase the granularity of the trajectory but optimize for token conservation due to the long-range of many CALVIN trajectories (upwards of 50 - 100 points).

\subsubsection{Spatial Relationship Identification}
In order to demonstrate the efficacy of the Spatial Prefix-Prompting mechanism beyond just 3D trajectory classification, we also run experiments with Llama-2-7B on the entire test set of the SpartQA dataset [\citenum{mirzaee-etal-2021-spartqa}] (of size 510 instances), a textual QA benchmark for deeper spatial reasoning questions of four types: find relation (FR), find blocks (FB), choose object (CO) (see Fig. ~\ref{fig:tasks}). 

\subsection{Metrics and Models}
As all of the tasks are classification / labeling tasks, we opt for traditional classification metrics, i.e. accuracy and F1-score, to reflect the balanced metrics of precision and recall. For the direction labeling, we also analyze the average number of direction misclassifications / errors per sequence, normalized by sequence length, calling this metric Err \# - this is effectively how many of the directions in a single sequence are erroneously predicted. Human evaluators evaluate the ChatGPT responses for correctness while heuristics (when the answer appears in the first or last line) are used for SpartQA. We use three models for our experiments: the first two are ChatGPT 3.5 and 4 [\citenum{openai2023gpt4}] and the third is Llama-7B [\citenum{touvron2023llama}] for SpartQA, chosen due to the quicker evaluation time for the size of the test dataset. We selected these models for their common use by the general public and quicker outputs.

\subsection{Results}
\begin{table}
  \caption{2D Direction and Shape Labeling performance on short, long and floating-point sequences}
  \label{sample-table}
  \centering
  \begin{tabular}{lllllllll}
    \toprule
    \multicolumn{1}{c}{} &
    \multicolumn{6}{c}{Direction Labeling} &
    \multicolumn{2}{c}{Shape Labeling}\\
    \cmidrule(r){2-7} 
    \cmidrule(r){8-9} 
    & \multicolumn{2}{c}{Integer (short)} &  \multicolumn{2}{c}{Float (short)} & \multicolumn{2}{c}{Integer (long)} & \multicolumn{1}{c}{Integer} & \multicolumn{1}{c}{Float} \\
    \cmidrule(r){2-3} 
    \cmidrule(r){4-5} 
    \cmidrule(r){6-7} 
    \cmidrule(r){8-8} 
    \cmidrule(r){9-9} 
    LLM     & Acc. ($\uparrow$) & Err. \# ($\downarrow$) & Acc. ($\uparrow$) & Err. \# ($\downarrow$) & Acc. ($\uparrow$) & Err. \# ($\downarrow$) & Acc. ($\uparrow$) & Acc. ($\uparrow$) \\
    \midrule \midrule
    ChatGPT-3.5 & 0.50  &  0.15  &  0.50 & 0.25  &  0.00 & 0.71 &  0.31 & 0.23 \\
    ChatGPT-4 & \textbf{1.00}  & \textbf{0.00}  & \textbf{1.00} & \textbf{0.00}  &  \textbf{0.60}  & \textbf{0.13} & \textbf{0.46} & \textbf{0.46}  \\
    \bottomrule
  \end{tabular}
  \label{table:2d}
\end{table}

Our main results across the three tasks are given in Table 1, Table 2 and Table 3, and the main findings are as follows.

\textbf{LLMs perform acceptable few-shot identification of directions   }
As seen in Table ~\ref{table:2d}, ChatGPT-3.5 and 4 succeed in achieving at least 50\% classification rates, with better performance (Err. \# <25) on shorter trajectories (len>5) and a neutral effect from the integer vs. floating point. We also see that parameter size and extensive training data likely play a huge role, with ChatGPT-4 hitting perfect classification for short trajectories and impressive performance (60\%) for long trajectories (len>35). The models struggle with Shape Labeling however, unsurprisingly from the lack of data in zero-shot evaluation. 

\begin{table}
  \caption{3D Trajectory Labeling performance for ChatGPT-3.5 \& 4 on CALVIN \& CALVIN-Cleaned} 
  \label{sample-table}
  \centering
  \begin{tabular}{llllll}
    \toprule 
    \multicolumn{2}{c}{} &
    \multicolumn{2}{c}{ChatGPT-3.5} &
    \multicolumn{2}{c}{ChatGPT-4} \\
    \cmidrule(r){1-2} 
    \cmidrule(r){3-4} 
    \cmidrule(r){5-6} 
    Dataset &  Method  & F1 ($\uparrow$) & Acc. ($\uparrow$) & F1 ($\uparrow$) & Acc. ($\uparrow$) \\
    \midrule \midrule
    \multirow{4}{*}{CALVIN} & Zero-shot & 0.19  & 0.26  &  0.19  &  0.27    \\
    & In-context & 0.17  &  0.33 & \textbf{0.63}  & \textbf{0.63}      \\
    & CoT &  0.28  &  0.36 & 0.33  & 0.37        \\
    & SPP &  \textbf{0.32} &  \textbf{0.43} & 0.55  & 0.60        \\
    \cmidrule(r){1-6}
    \multirow{4}{*}{CALVIN-Cleaned} & Zero-shot & 0.24  &  0.30 & 0.42  & 0.47     \\
    & In-context & 0.16  &  0.34 & 0.62  & 0.67    \\
    & CoT & 0.14 &  0.26 & 0.73  & 0.73       \\
    & SPP &  \textbf{0.38}  &  \textbf{0.43} & \textbf{0.80}  & \textbf{0.80}      \\
    \bottomrule
  \end{tabular}
  \label{table:3d}
\end{table}

\textbf{LLMs demonstrate poorer capabilities on more complex 3D trajectories  }
We find that, as seen in Table ~\ref{table:3d}, LLMs achieve subpar performance when compared with the 2D scenario, especially on the raw data from the CALVIN benchmark, with the highest F1-score capped at 63\%. In their current form, it is improbable that such LLMs can be used for robotic trajectory classification. A possible cause for the disparity between the CALVIN and CALVIN-Cleaned datasets could be the higher degree of irregularity in the CALVIN dataset, since as demonstrated in [\citenum{mirchandani2023large}], such models excel in mimicking more repetitive patterns (e.g. sinusoidal graphs). Another factor could be that LLMs connect spatial patterns to pre-trained semantic concepts in the CALVIN-Cleaned dataset - for example, the cleaned "lifting" trajectory illustrates only a downward and upward motion in the z-dimension, matching a fundamental understanding perhaps baked in from pretraining, whereas the extraneous datapoints might muddle such comprehension.

\begin{figure}
  \centering
  \includegraphics[width=8cm]{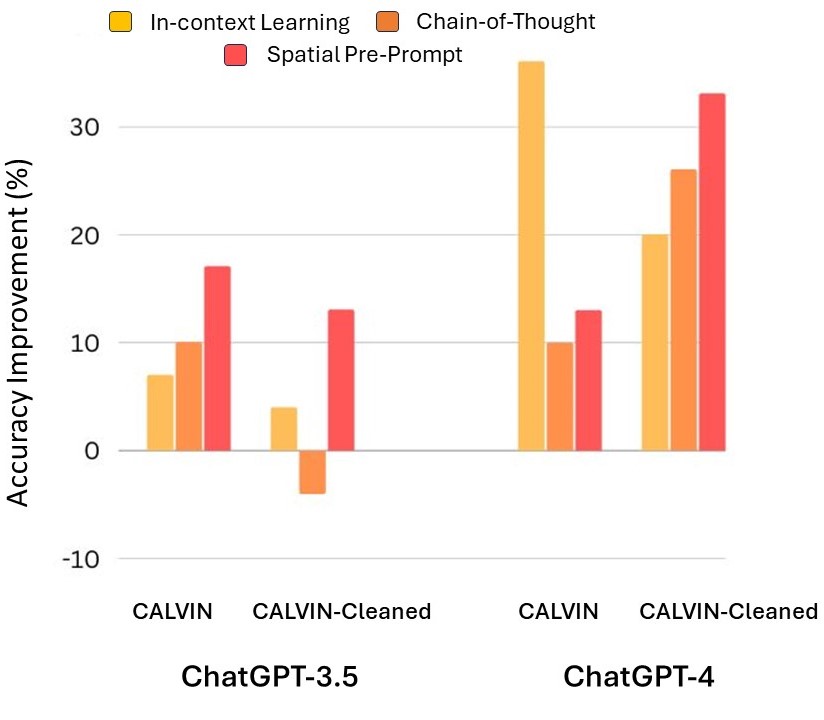}
  \caption{Accuracy Improvements (\%) for In-context Learning, Chain-of-Thought and Spatial Prefix-Prompting on both the CALVIN and CALVIN-Cleaned datasets, for ChatGPT-3.5 and ChatGPT-4. As we can see, overall the accuracy gains for ChatGPT-4 are higher.}
  \label{fig:acc_gain}  
\end{figure}

\textbf{CoT reveals a reduction in spatial reasoning performance }
Table ~\ref{table:3d} and Fig. ~\ref{fig:acc_gain} conveys the volatility in the performance of CoT, revealing either losses or marginal performance gains (mostly bounded at 11\%) compared to In-context Learning (ICL), and even the gains are much lower than other tasks [\citenum{wei2023chainofthought}]. 
A potential reason for the diminishing returns in this application is that the CoT reasoning steps are fairly dependent on the examples chosen. We have qualitatively seen examples in which the model understands a single example in the context of its action (e.g. since an object is performing the action "lift", its z-coordinate decreases), but then witnesses a slight decrease in the z-coordinate of a "slide" trajectory due to noisiness and concludes that it belongs to the "lift" category. 

\textbf{LLMs seem to enable knowledge transfer from simple to more complex tasks}
From Tables ~\ref{table:3d} and ~\ref{table:spartqa}, we observe that Spatial Prefix-Prompting (SPP) often surpasses CoT and ICL, particularly on the CALVIN-Cleaned dataset and the "Find Relationship" (FR) and "Choose Object" (CO) questions in the SpartQA dataset [\citenum{mirzaee-etal-2021-spartqa}]. This outcome hints that SPP might perhaps be better suited to scenarios in which the labels themselves hold morphological meaning, permitting the model to expand upon its pretrained knowledgebase (e.g. in the FR and CO questions, the labels refer to directional relationships like "above" or qualitative adjectives "medium black square").
Furthermore, it has previously been corroborated that, taking a Bayesian lens, ICL operates by helping the model to locate latent concepts that it learned during pretraining [\citenum{xie2022explanation}, \citenum{min2022rethinking}, \citenum{razeghi-etal-2022-impact}], i.e. if terms in a particular instance are exposed many times in the pretraining data, the model is likely to know better about the distribution of the inputs. It can be that SPP operates similarly, with a simple spatial question (e.g. direction identification) prodding the model to draw upon a more fundamental mechanism that it has been trained on (e.g. calculating numerical differences between coordinates to designate directions), in order to solve more complex questions that may use an analogous thought-process.

\begin{table}
  \caption{Llama 2 7B performance on the SpartQA Test Dataset, split by subtypes FR, FB, CO, YN}
  \label{sample-table}
  \centering
  \begin{tabular}{lllllll}
    \toprule
    \multicolumn{2}{c}{} &
    \multicolumn{1}{c}{FR} &
    \multicolumn{1}{c}{FB} &
    \multicolumn{1}{c}{CO} &
    \multicolumn{1}{c}{YN} &
    \multicolumn{1}{c}{Overall}\\
    \cmidrule(r){3-3}
    \cmidrule(r){4-4} 
    \cmidrule(r){5-5}
    \cmidrule(r){6-6}
    \cmidrule(r){7-7}
    LLM & Method & Acc. ($\uparrow$) & Acc. ($\uparrow$) & Acc. ($\uparrow$) & Acc. ($\uparrow$) & Acc. ($\uparrow$) \\
    \midrule \midrule
    \multirow{2}{*}{Llama 2 7B} & Zero-shot & 0.14 & 0.40 &  0.24  &  0.39  &  0.32 \\
    & CoT & 0.21 & \textbf{0.47}  &  0.16  &  \textbf{0.48}  &  0.36  \\
    & SPP & \textbf{0.42} & 0.44  &  \textbf{0.42}  &  0.40  &  \textbf{0.41}  \\
    \bottomrule
  \end{tabular}
  \label{table:spartqa}
\end{table}

\section{Conclusion}

We examined the performance of LLMs including ChatGPT 3.5, 4 and Llama 2 7B on a variety of spatial tasks, namely 2D direction and path labeling, 3D trajectory labeling and abstract relationship identification. We show that the selected models exhibit acceptable performance on 2D direction labeling but flounder to a greater deal on 3D trajectory labeling. We speculate on possible causes, settling on the likelihood that the irregularity of the trajectories makes classification more onerous. We also hypothesize that the brittleness of Chain-of-Thought prompting's reliance on specific examples influences its diminished yield in noisy scenarios. Finally, we propose a technique called Spatial Prefix-Prompting that first inquires a simple, related question in order to better answer more complex spatial queries. Our work could have implications in a multitude of other domains than just higher-dimensional numerical data, such as multi-variable financial trend forecasting or aggregate health data analysis. Future work includes evaluation on a larger robotic dataset, extension to other spatial tasks (e.g. segmenting trajectories), and assessment of other LLMs like PaLM \citenum{chowdhery2022palm}. Overall, we establish that the domain of spatial reasoning, especially with regards to numerical data, is an underexplored realm ripe for more research.

\bibliography{sample}

\medskip

\small

\end{document}